\documentclass[conference]{IEEEtran}
\IEEEoverridecommandlockouts
\usepackage{cite}
\usepackage{amsmath,amssymb,amsfonts}
\usepackage{algorithmic}
\usepackage{graphicx}
\usepackage{textcomp}
\usepackage{xcolor}
\usepackage{multirow}
\usepackage{makecell}
\usepackage{bm}
\usepackage{adjustbox}
\usepackage{hyperref}

\def\BibTeX{{\rm B\kern-.05em{\sc i\kern-.025em b}\kern-.08em
    T\kern-.1667em\lower.7ex\hbox{E}\kern-.125emX}}
\begin{document}

\title{No-Reference Point Cloud Quality Assessment via Weighted Patch Quality Prediction\\
% {\footnotesize \textsuperscript{*}Note: Sub-titles are not captured in Xplore and should not be used}
% \thanks{Identify applicable funding agency here. If none, delete this.}
}

\author{
\IEEEauthorblockN{Jun Cheng}
\IEEEauthorblockA{Shenzhen University\\
chengjun2020@email.szu.edu.cn
}

\and
\IEEEauthorblockN{Honglei Su*}
\IEEEauthorblockA{Qingdao University\\
suhonglei@qdu.edu.cn
}
% \authornote{*Corresponding authors}
% (Corresponding Author)

\and
\IEEEauthorblockN{Jari Korhonen}
\IEEEauthorblockA{University of Aberdeen\\
 jari.t.korhonen@ieee.org
}
\thanks{DOI reference number: 10.18293/SEKE2023-185}
}

\maketitle

\begin{abstract}
With the rapid development of 3D vision applications based on point clouds, point cloud quality assessment (PCQA) is becoming an important research topic. However, the prior PCQA methods ignore the effect of local quality variance across different areas of the point cloud. To take an advantage of the quality distribution imbalance, we propose a no-reference point cloud quality assessment (NR-PCQA) method with local area correlation analysis capability, denoted as COPP-Net. More specifically, we split a point cloud into patches, generate texture and structure features for each patch, and fuse them into patch features to predict patch quality. Then, we gather the features of all the patches of a point cloud for correlation analysis, to obtain the correlation weights. Finally, the predicted qualities and correlation weights for all the patches are used to derive the final quality score. Experimental results show that our method outperforms the state-of-the-art benchmark NR-PCQA methods. The source code for the proposed COPP-Net can be found at \href{https://github.com/philox12358/COPP-Net}{https://github.com/philox12358/COPP-Net}.

\end{abstract}


\section{Introduction}
A 3D point cloud is a large and dense collection of sampled points with spatial coordinates and attributes, obtained by 3D scanning technology like lidar. Each point in a point cloud contains a geometric attribute, i.e. 3D space coordinates, and other attributes, such as color, reflectivity, opacity, etc., represented by feature vectors. As point clouds directly represent the 3D world, they are widely used in automatic driving, industrial robots, cultural heritage protection, geographic mapping, and other fields.

Similar to image and video, point clouds can be distorted due to many factors during the collection and transmission process. However, point clouds suffer from more complex distortions due to their data format. Therefore, PCQA is more challenging than traditional image quality assessment. Accurate PCQA is critical for providing high-quality point clouds for various purposes. Like in traditional image and video quality assessment, PCQA methods can also be divided in full-reference (FR), reduced-reference (RR), and no-reference (NR) methods, according to the availability of a reference point cloud.

Several FR-PCQA methods have been proposed recently~\cite{mekuria2016evaluation}~\cite{mekuria2017performance}~\cite{meynet2020pcqm}~\cite{wang2004image}~\cite{yang2020inferring}~\cite{sheikh2006image}~\cite{liu2022perceptual}, and they have reached relatively good performance already. However, in many real-life application scenarios, there is no pristine reference point cloud available, and NR-PCQA is therefore a very important research topic at the moment. Regarding the recent development of image quality assessment (IQA), the most accurate NR-IQA methods are based on deep convolutional neural networks (CNN). Unfortunately, training of deep neural networks (DNN) requires a large amount of training samples, and the currently available PCQA databases are mostly small, and each point cloud often contains millions of points. Advanced point convolution neural networks directly processing such a large number of points have a very high computational load, limiting the use of deep neural networks for NR-PCQA.

\begin{figure}[t]             % correlation map
\begin{center}
    \includegraphics[width=1.0\linewidth]{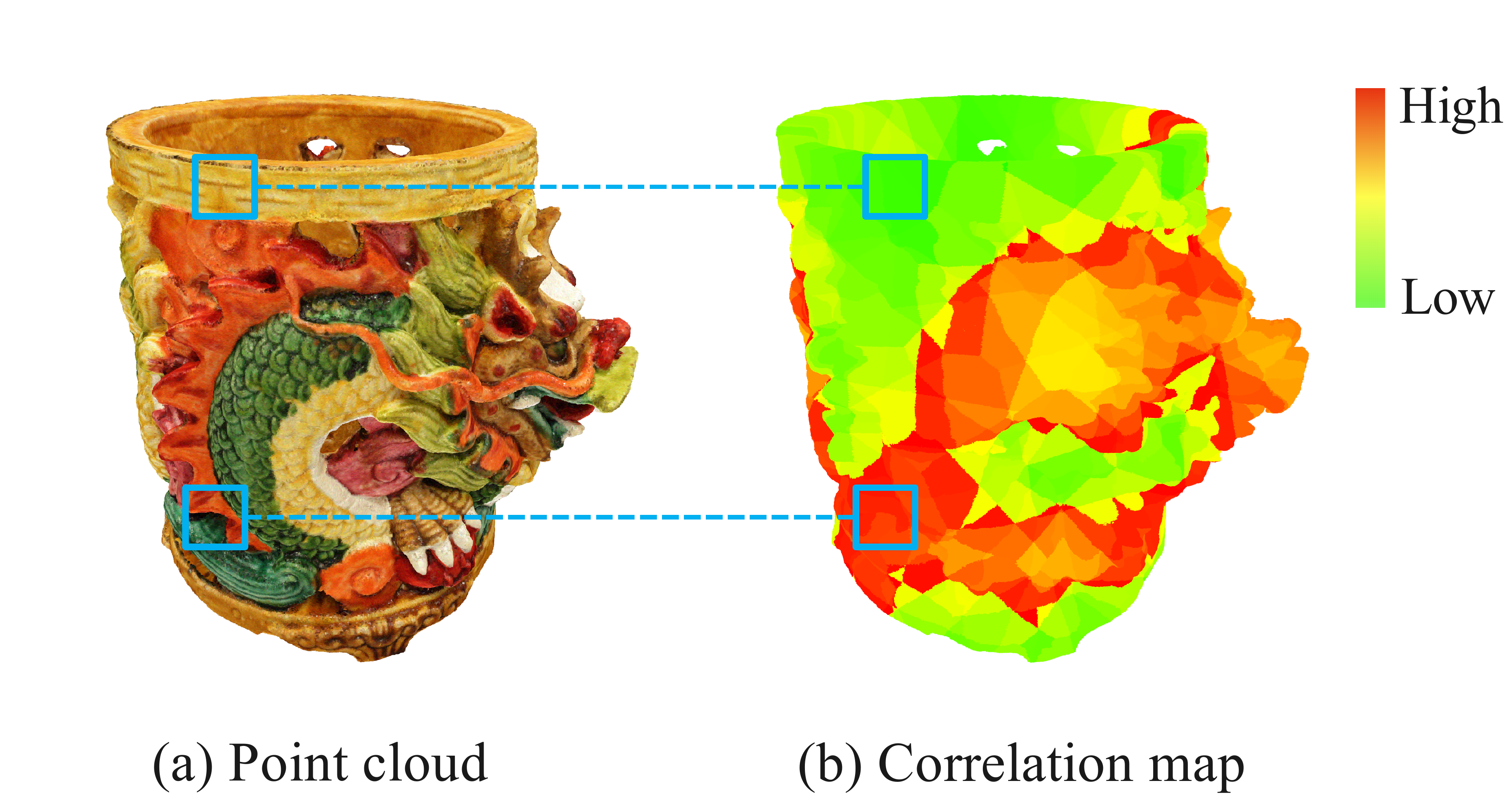}
\end{center}
\vspace{-0.1cm}
   \caption{Visualization of correlation. (a) shows that the structure and texture of different parts of the point cloud are very different; (b) shows that the correlation between the predicted patch quality and the overall quality is inconsistent. Patches with different correlations should have different weights for predicting the overall quality.}
\label{fig:correlation}
\end{figure}

\begin{figure*}[t]       % COPP_0-main
\begin{center}
    \includegraphics[width=1.0\linewidth]{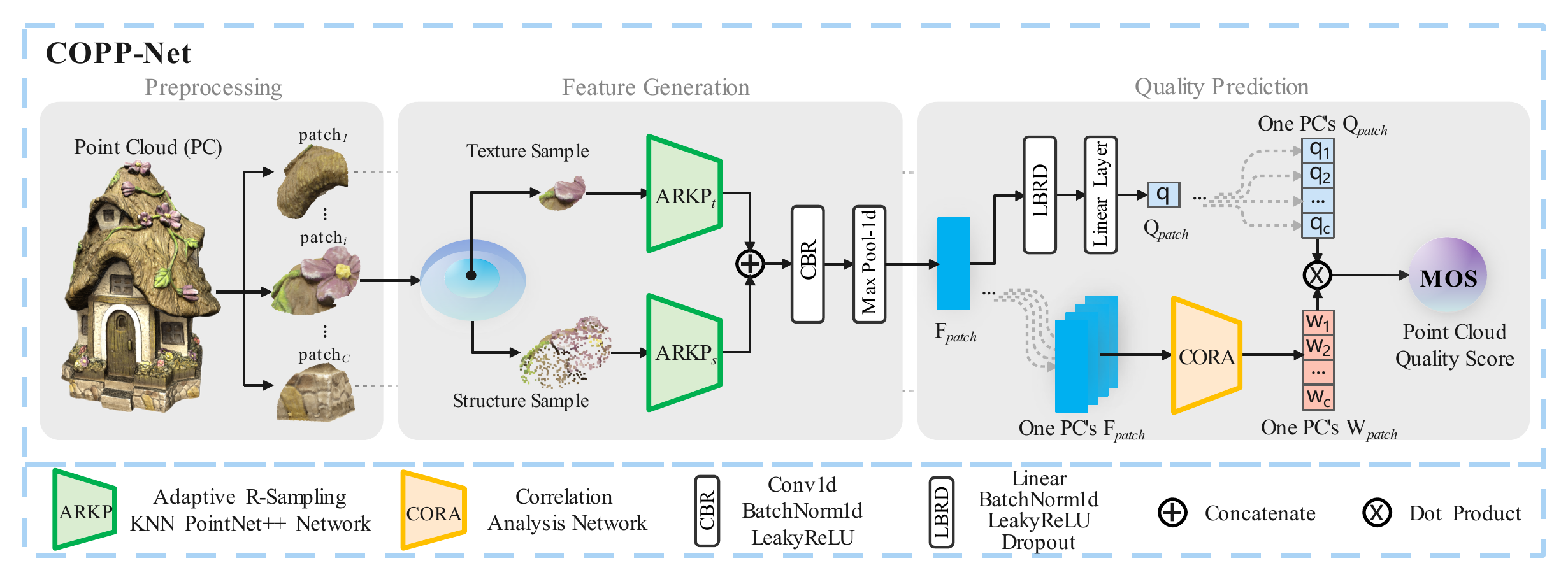}
\end{center}
\vspace{-0.3cm}
   \caption{The overall architecture of COPP-Net with preprocessing, feature generation and quality score prediction modules. In the preprocessing module, FPS and KNN algorithms are used to obtain the specified number of patches with specified size. The feature generation module in the middle contains two parallel ARKP networks, named ARKP$_{t}$ and ARKP$_{s}$. Last, the quality score prediction module uses the patch features (F$_{patch}$) to generate the patch quality scores (Q$_{patch}$) and weights (W$_{patch}$) for each patch, and then it uses weighted averaging to derive the quality score for the point cloud.}
\label{fig:model structure}
\end{figure*}

Although some attempts have been made to use DNNs for NR-PCQA~\cite{liu2021pqa}~\cite{zhang2022no}~\cite{yang2022no}, these methods usually obtain projected 2D images from the whole point cloud or a few features of the whole point cloud as the input of the network through preprocessing, rather than directly using the point cloud. ResSCNN~\cite{liu2022point} directly uses the entire point cloud as input, but sparse convolution is used to save computation, which reduces the computational accuracy. NR-CNN-3D-PC~\cite{chetouani2021deep} divides the point cloud into many patches, but does not consider the impact of different patches on the overall quality of the point cloud. At the same time, compared with the most accurate FR-PCQA method, the proposed models still perform poorly. 
So far, we have not identified any stable and accurate method of NR-PCQA task that uses point cloud as the input of DNN model directly and takes into account the local correlation of point cloud.

Due to the development of new methods for deep learning in the recent years, such as PointNet++~\cite{qi2017pointnet++} and transformers~\cite{vaswani2017attention}, there have been remarkable progress in processing point clouds with fewer points recently. On the other hand, in many practical use cases large 3D point clouds need to be observed from different sides and viewing angles multiple times through rotation and scaling. Moreover, point cloud usually has complex structure and rich texture, and the visual appearance of different regions can be very different. Therefore, the correlation between the local quality and the overall quality varies strongly between different areas of the a point cloud.  Based on these considerations and the previous work, we propose a patch-based NR-PCQA method that generates features representing the structure and texture of each patch. Then, point cloud quality is predicted by using appropriate weights derived from the correlations between patch quality and overall quality. 

It is worth noting that correlation in our method is different from saliency. Even though the correlations may be related to saliency, our method does not aim to calculate the saliency map of the point cloud. Specifically, we first divide a point cloud into several patches, representing different parts of the point cloud. In the proposed COPP-Net model, we propose Adaptive R-Sampling KNN PointNet++ (ARKP) network, based on PointNet++ architecture~\cite{qi2017pointnet++}, to generate texture and structure features from patches generated from the point cloud. These features are used to predict local quality scores for the patches, and they are also used as input to the proposed Quality Correlation Analysis (CORA) network, predicting the weights for each patch. Finally, the quality scores for each patch are multiplied by their relative weights to predict the overall quality score through a weighted average.

\section{Related work}
% In PCQA, traditional methods rely on handmade features. Recently, learning based methods mainly include projection based schemes that directly use point clouds. These programmes are outlined here.

\subsection{Full-Reference metrics}
For PCQA, several FR metrics were firstly developed to be used for assessing performance of point cloud compression. The classic methods are point-based FR metrics proposed by MPEG, such as p2point~\cite{mekuria2016evaluation} and p2plane~\cite{tian2017geometric}. Their computational complexity is low, but on the other hand, their accuracy is limited and they are prone to instability when complex distortion types are concerned. In~\cite{alexiou2018point}, a point-based metric based on angular similarity was proposed, and~\cite{meynet2019pc} proposed a series of methods using local luminance patterns, local binary patterns and multi-distance approach. In~\cite{javaheri2020generalized}, a generalized Hausdorff distance based quality metric for geometric point cloud distortions was proposed, and
~\cite{meynet2019pc} used the local curvature statistics to evaluate the quality of point clouds. Later, inspired by SSIM~\cite{wang2004image}, several metrics were proposed in~\cite{yang2020inferring}~\cite{liu2022perceptual}. They considered the structural characteristics of the point cloud and achieved a good performance for FR-PCQA.

\subsection{No-Reference metrics}
Due to the limited availability of point cloud quality databases, exploration of NR-PCQA started relatively late. Since NR-IQA methods have matured~\cite{liu2021pqa}, it has been proposed to project point clouds to 2D pictures from different angles, and then use traditional IQA methods or CNN to predict point cloud quality indirectly from the 2D images. In NR-3DQA~\cite{zhang2022no}, 3D point clouds are projected into feature domains based on color and geometry, and the quality score is obtained using support vector machine (SVM) regression. IT-PCQA~\cite{yang2022no} using the rich prior knowledge of natural images to build a bridge between 2D and 3D perception for quality assessment via transfer learning. 
ResSCNN~\cite{liu2022point} uses 3D sparse convolution for efficient computations. However, in the process of sparse convolution, the local accuracy of the point cloud data will be reduced, resulting in partial information loss. Therefore, the performance of the method is not stable on point clouds with different types of distortions. 

In~\cite{chetouani2021deep}, point clouds are split into multiple local patches, and low level patch-wise features (e.g., geometric distance, local curvature) are used to train a CNN. However, this method does not take into account the quality inconsistency between different areas of the point cloud. To tackle this challenge, we propose a model that learns the weights for different parts of the point cloud to balance the inconsistencies in local quality.

\begin{figure}[t]       % ARKP
\begin{center}
    \includegraphics[width=1\linewidth]{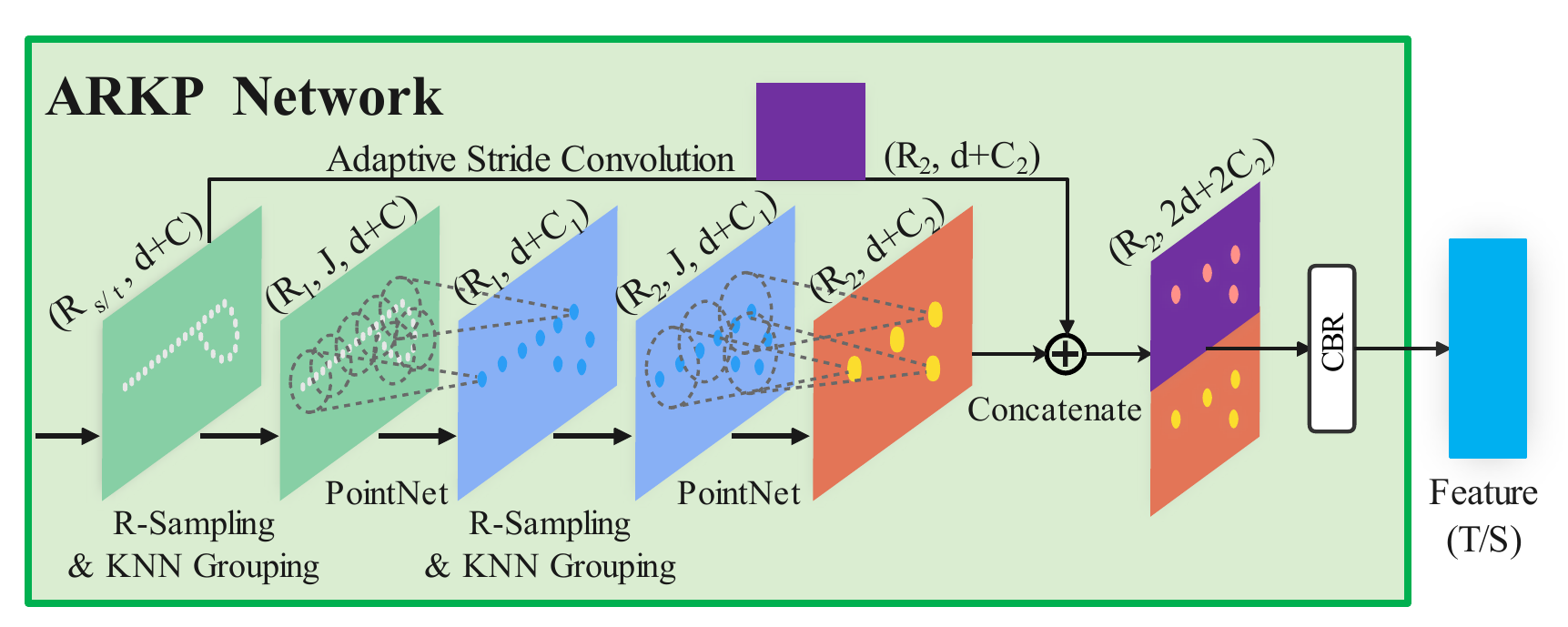}
\end{center}
\vspace{-0.1cm}
   \caption{Detailed structure of ARKP network.}
\label{fig:ARKP}
\end{figure}

\section{Our Approach}
% \subsection{Overview}
% Most of the assets in the common point cloud data set contain millions of points, and such a large number will consume huge computing resources when generating features. When observing a point cloud, people usually rotate to different viewpoints to observe local performance from different angles, and finally synthesize the quality score of the whole point cloud. In this process, different parts of the local have different effects on the overall quality. Based on the above considerations, our method is to generat multiple local patches from the point cloud and score each local patch separately. Then, the correlation between the learned local patch and the overall quality is used as the weight to obtain the prediction quality of the entire point cloud. This method is similar to the human visual scoring mechanism, and reduces the amount of computation.

% As shown in the figure, the difference in quality scores of different local patches of each point cloud is obvious, and the pooling strategy using correlation weights is more consistent with human vision than the average pooling strategy.

\subsection{Point Cloud Preprocessing Module}
A point cloud instance is a set $\mathbf{P}$ that contains a total of $N$ points, $\mathbf{P} = \{P_{i}\ |\ i=1,2,...,N\}$, located on the surface of the object of interest. Note that $P_{i}=(x_{i}, y_{i}, z_{i}, r_{i}, g_{i}, b_{i})$ defines the $i$-th point of this point cloud, and the spatial coordinates denoted as $(x,y,z)$ and the color information denoted as $(r,g,b)$. Thus, a 3D point cloud object can be represented by a matrix of size $N \times 6$. Generally, $N$ is in millions. Processing that many points will consume a lot of computing resources. Considering that different parts of the point cloud have different quality correlations, we divide the point cloud into multiple patches through preprocessing.

For preprocessing, we first normalize $(x,y,z)$ into a sphere with a radius of 1000, then we use the farthest point sampling (FPS)~\cite{qi2017pointnet++} algorithm to obtain $C$ center points, and finally we use the k-nearest neighbor (KNN)~\cite{abeywickrama2016k} algorithm to sample the nearest $K$ points to each center point to form the patches. For each patch, we center the $(x,y,z)$ coordinates and the $(r,g,b)$ color information, respectively. As a result, we sample $C$ patches from a point cloud, and each patch contains $K$ points. These patches are then used as input to the patch feature generation module.

\begin{figure}[t]       % pc_patch_mos
\begin{center}
    \includegraphics[width=0.95\linewidth]{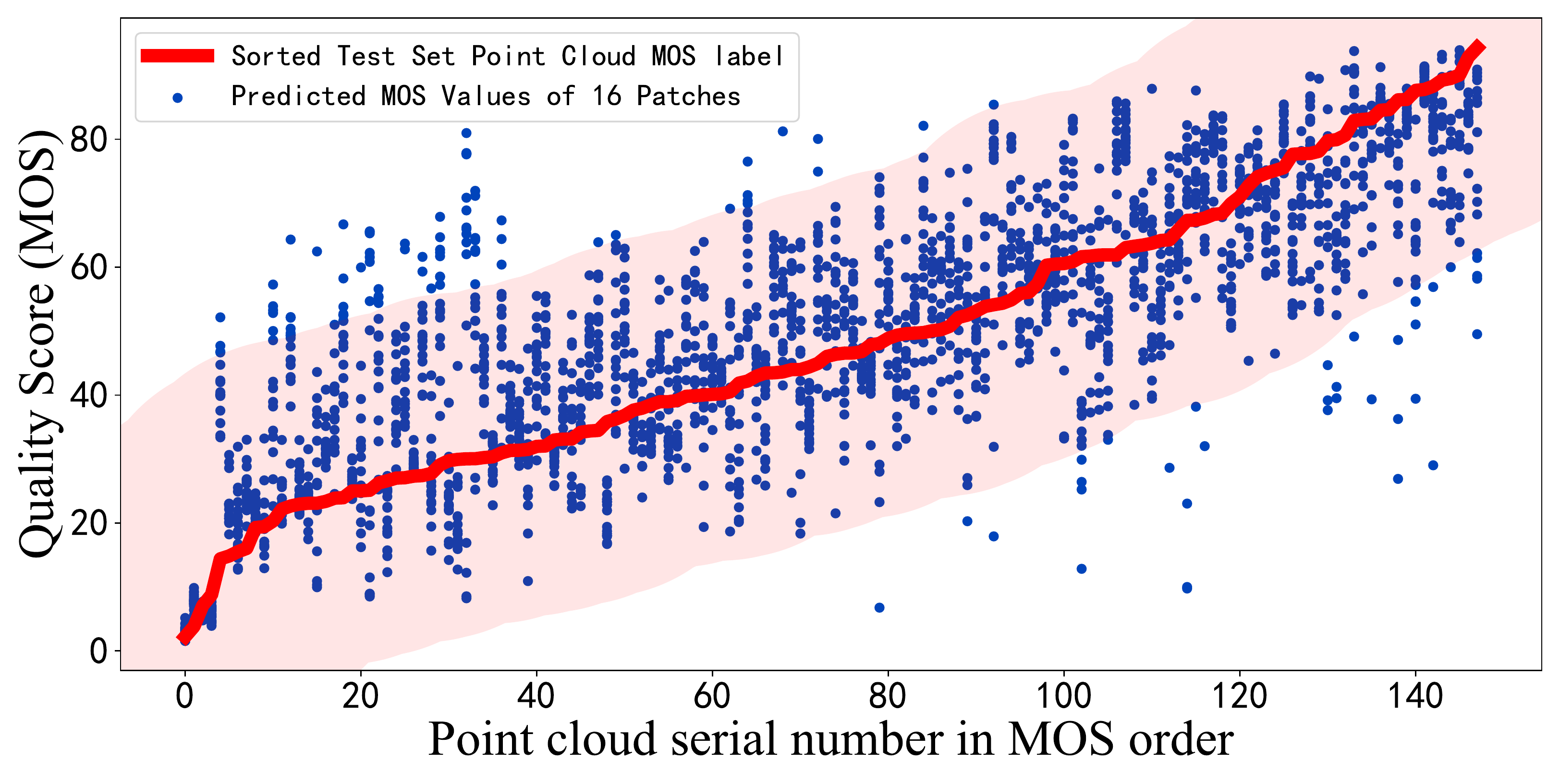}
\end{center}
\vspace{-0.2cm}
   \caption{Overall point cloud MOS (red) and predicted Q$_{patch}$ for patches (blue) on WPC database.}
\label{fig:pc_patch_mos}
\end{figure}

\subsection{Patch Feature Generation Module}
The feature generation module will generate features for texture and structure information. Texture information is very sensitive to downsampling, because it requires fine and continuous dense data representation. Fortunately, continuous regions with similar quality scores tend to share similar texture distortion characteristics. Therefore, we can generate texture features from small regions without losing the accuracy of the representation. For structural information, rough and sparse information is sufficient to represent complex spatial structures, so structural information is not sensitive to downsampling. Based on these observations, we follow different strategies to generate texture and structure features.

Specifically, the module is composed of local texture feature generation network (ARKP$_{t}$) and 3D structure feature generation network (ARKP$_{s}$). Both networks are based on ARKP architecture, and they are used to generate texture and structure features, respectively. The details are as follows:

\textbf{Texture feature generator ARKP$_{t}$.} 
In a patch with $K$ points, we use KNN algorithm to sample locally adjacent and continuous $R_{t}$ points to be used as input to ARKP$_{t}$ for obtaining the texture features of the patch.

\textbf{Structure feature generator ARKP$_{s}$.} 
In a patch with $K$ points, we use random sampling to sample a relatively small number of $R_{s}$ points within the range of the patch to be used as input to ARKP$_{s}$ for obtaining the structure features of the patch.

Then, we concatenate the features from the two networks and send it through a MaxPool layer to obtain the final patch feature vectors (F$_{patch}$). F$_{patch}$ will be used to predict the patch quality score (Q$_{patch}$). It is also used by the CORA network to obtain correlation weight (W$_{patch}$). We will present a detailed description of the ARKP network below.

\subsection{ARKP Network}
We propose the ARKP network to generate features more accurately.
As depicted in Figure~\ref{fig:ARKP}, the ARKP architecture is based on the single-scale grouping (SSG) version of PointNet++~\cite{qi2017pointnet++}.

In PointNet++, the hierarchical structure is composed of a number of specific abstraction levels, each of which is made up of three key layers: the Sampling layer, Grouping layer, and PointNet layer~\cite{qi2017pointnet++}. The Sampling layer defines the centroids of local regions using FPS algorithm, and the Grouping layer constructs local regions by performing a ball query. The PointNet layer encodes the local region into feature vectors.

The ARKP network differs from the SSG version of PointNet++ in the following aspects:
\begin{itemize}
    \item To improve the overall information generation ability, we added Adaptive Stride Convolution before and after the baseline structure.
    \item To reduce computational overhead, we use Random Sampling (R-Sampling) instead of FPS in the sampling layer.
    \item To improve stability in PCQA tasks, we use KNN to select neighboring points in the Grouping layer.
\end{itemize}
Our ablation experiments demonstrate the effectiveness of the proposed modifications.

\subsection{Point Cloud Quality Regression Module}
Figure~\ref{fig:model structure} shows a detailed overview of the steps to predict the point cloud quality score from the F$_{patch}$. Using F$_{patch}$ as input, we apply a regression head comprising two linearlayers, batch normalization layer, and leaky ReLU layer to predict Q$_{patch}$. In the training phase, we assign the overall point cloud quality score as ground truth quality score for all the patches. Mean squared error (MSE) is used as loss function for training. If the quality score of a point cloud is calculated by averaging the values of Q$_{patch}$, high prediction accuracy seems to be achieved.

However, as shown in Figure~\ref{fig:pc_patch_mos}, the quality scores for individual patches tends to be scattered, since different areas of the point cloud have different quality levels. To reduce the impact of outlier patches, we use a network (CORA) to analyze the dispersion of quality of each patch.

\begin{figure}[t]             % CORA
\begin{center}
    \includegraphics[width=1.0\linewidth]{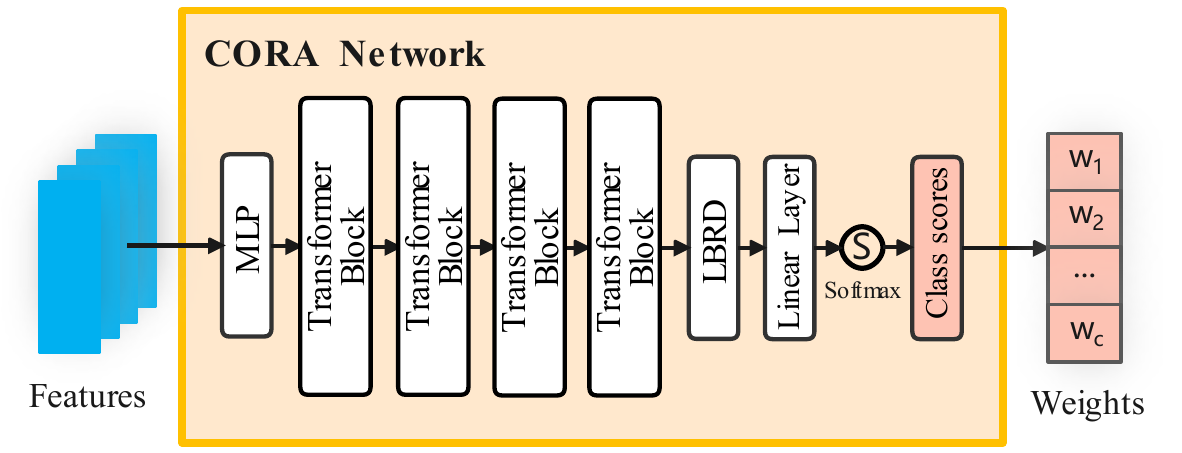}
\end{center}
\vspace{-0.1cm}
   \caption{Detailed structure of CORA network.}
\label{fig:CORA}
\end{figure}

\subsection{CORA Network}
The CORA network is designed to estimate the correlation between the patch quality and the overall point cloud quality by predicting correlation labels. As illustrated in Figure~\ref{fig:CORA}, we concatenate all F$_{patch}$ of a single point cloud to form the input to the CORA network. Then, we apply a multilayer perceptron (MLP) with two linear layers, followed by four transformer blocks and two additional linear layers to predict the correlations. The middle hidden layer dimension is set to 512. Finally, SoftMax operation is applied to the output to compute the correlation-based W$_{patch}$.

Specifically, we obtain all Q$_{patches}$ of a point cloud and rank them in ascending order based on the absolute difference between each Q$_{patch}$ and the ground truth quality score of the point cloud.
Then divide them into three correlation classification labels, i.e., strong correlation, average correlation, and weak correlation. The goal of the CORA network is to predict the correlation classifications accurately. We use cross-entropy loss to train the CORA network.

In the final phase of computing the predicted overall quality score, we use correlation weight pooling. Specifically, we use Q$_{patches}$ and W$_{patches}$ from the CORA network, and then compute the weighted average of the quality scores to obtain the final quality score of the point cloud. Formally, the correlation weight pooling can be expressed as:

\begin{eqnarray}
{\rm Q}_{PC} = \frac{\sum_{i=1}^{C}{{\rm W}_{i}{\rm Q}_{i}}}{\sum_{i=1}^{C}{\rm W}_{i}},
\end{eqnarray}
where Q$_{PC}$ is the quality score of the point cloud, $C$ is the number of patches generated from each point cloud, W$_{i}$ is the correlation weight for $i$-th patch, and Q$_{i}$ is the predicted quality score for $i$-th patch.

\begin{table}[t]
\begin{center}
\caption{Results on WPC database using several NR methods. We report the PLCC, SRCC and RMSE indicators.} 
\label{tab:wpc_contrast}
\setlength{\tabcolsep}{3.5mm}{
\begin{tabular}{|c|l|c|c|c|}
  \hline
  Type & Method & PLCC & SRCC &RMSE  \\
  \hline
  \multirow{5}{*}{NR}     
    & NIQE~\cite{mittal2012making}     &0.3957 &0.3887 &22.55 \\
    & BRISQUE~\cite{mittal2012no}      &0.4176 &0.3781 &22.54 \\
    & PQA-Net~\cite{liu2021pqa}        &0.7000 &0.6900 &15.18 \\
    & NR-3DQA~\cite{zhang2022no}       &0.6591 &0.6326 &16.45 \\
    & ResSCNN~\cite{liu2022point}      &0.4292 &0.4352 &23.27 \\
    & IT-PCQA~\cite{yang2022no}        &0.5609 &0.5683 &19.07 \\
    & Our COPP-Net     &\textbf{0.9324}&\textbf{0.9251}&\textbf{8.10} \\
  \hline
\end{tabular}}
\end{center}
\end{table}

\begin{table*}[t]
\begin{center}
\scriptsize
\caption{Comparison of our COPP-Net with the FR method on the test source contents in WPC database. In all cases, the proposed method performs better than the FR methods. The results for the FR methods are from~\cite{liu2022perceptual}.} 
\label{tab:combat FR}
\adjustbox{max width=1.0\linewidth,width=1.0\linewidth,center=\linewidth}{
\begin{tabular}{|c|l|ccc|ccc|ccc|ccc|}
	\hline
	\multirow{2}{*}{Type} &\multirow{2}{*}{Method} & \multicolumn{3}{c|}{Banana} & \multicolumn{3}{c|}{Cauliflower} & \multicolumn{3}{c|}{Mushroom}& \multicolumn{3}{c|}{Pineapple}\\
	\cline{3-14}       
	 & & PLCC & SRCC & RMSE & PLCC & SRCC & RMSE & PLCC & SRCC & RMSE & PLCC & SRCC & RMSE\\
	\hline
	\multirow{7}{*}{FR} 
	& PSNR$_{p2po,M}$~\cite{mekuria2016evaluation} &0.7236&0.6471&14.98&0.4555&0.3501&19.95&0.6406&0.6396&17.02&0.4678&0.3777&18.04 \\
    & PSNR$_{Y}$~\cite{mekuria2017performance}    &0.7560&0.6785&14.20&0.6332&0.5927&17.35&0.8022&0.6550&13.23&0.7466&0.7217&13.58 \\
    & PCQM~\cite{meynet2020pcqm}                &-0.7145&-0.7686&15.18&-0.7057&-0.6971&15.88&-0.8033&-0.7819&13.20&-0.7578&-0.7862&13.32 \\
    & SSIM$_{p}$~\cite{wang2004image}             &0.7156&0.7544&15.16&0.6515&0.5967&17.00&0.7994&0.7297&13.32&0.7214&0.7193&14.14 \\
    % & MS-SSIM$_{p}$~\cite{wang2003multiscale}     &0.7812&0.7968&13.54&0.6068&0.5730&17.82&0.8296&0.8535&12.38&0.7134&0.7105&14.3 \\
    & GraphSIM~\cite{yang2020inferring}         &0.5990&0.5300&17.38&0.6010&0.5520&17.91&0.7120&0.6730&15.56&0.0410&0.0310&20.20 \\
    & VIFP$_{p}$~\cite{sheikh2006image}           &0.7938&0.7956&13.20&0.6581&0.5820&16.88&0.8450&0.8658&11.85&0.7456&0.7285&13.61 \\
& IW-SSIM$_{p}$~\cite{liu2022perceptual}         &0.8724&0.8627&10.61&0.8578&0.8125&11.52&0.8697&0.8528&10.94&0.7817&0.7584&12.73 \\
 \hline
\multirow{5}{*}{NR}
    &PQA-Net~\cite{liu2021pqa}
    &0.5300&0.5200&18.87&0.7000&0.6900&15.27 &0.7700&0.7100&13.81&0.8700&0.8900&9.80 \\
    &NR-3DQA~\cite{zhang2022no}
    &0.6965&0.7008&16.13&0.5977&0.5744&18.42&0.7740&0.8297&15.17&0.6288&0.5915&15.90 \\
    &ResSCNN~\cite{liu2022point}
    &0.4484&0.4694&25.04&0.4113&0.3340&23.65&0.5259&0.5209&22.48&0.7383&0.6747&22.51 \\
    &IT-PCQA~\cite{yang2022no} 
    &0.7855&0.8377&13.58&0.7156&0.7146&15.20&0.2989&0.3097&22.05&0.2923&0.3702&20.32 \\
    & Our COPP-Net  
    &\textbf{0.9121}&\textbf{0.8878}&\textbf{9.07}&\textbf{0.9190}&\textbf{0.9246}&\textbf{9.54}&\textbf{0.9495}&\textbf{0.9272}&\textbf{7.72}&\textbf{0.9571}&\textbf{0.9478}&\textbf{6.02}
     \\
 	\hline
\end{tabular}}
\end{center}
\end{table*}

\section{Experiment}
In this section, we first describe the experimental setup, then we introduce the Waterloo Point Cloud Database (WPC), and compare the performance of our COPP-Net with the prior mainstream FR and NR methods. Then, we conduct an ablation study on the proposed COPP-Net. Finally, we test the performance of the model on other databases.

\subsection{Experiment Setups}
% With regard to patch information, in order to balance the computational load and network performance, 
In each of the practical experiments, we extracted $C$=16 patches from a point cloud, with $K$=14900 points in each patch. In the feature generation module, we used KNN algorithm to select $R_{t}$=8192 adjacent points around the center of a patch as input to ARKP$_{t}$, and random sampling to select $R_{s}$=1024 points as input to ARKP$_{s}$. 

The model was trained in two stages: in the first stage, we trained the feature generation and patch quality prediction part, and in the second stage, we trained the CORA network. 
%We used all the patches from the training set in a random order to train the patch quality prediction part. The overall quality score of the point cloud was assigned as ground truth to all the patches generated from the point cloud, and MSE was used as loss function. 

% In the second stage, we used fixed weights for the feature generator and patch quality prediction parts, obtained from the training process in the first stage. Then, we used the patch quality prediction model to get the predicted quality scores for all the patches in the training set, and ordered the predicted scores for each point cloud according to the difference between the predicted patch score and the ground truth overall point cloud score. The six patches with the smallest difference were assigned a label for strong correlation, the following five patches were assigned a label for average correlation, and the last five patches were classified as weakly correlated. These classifications were used as ground truth for training the CORA module. Cross entropy was used as loss function.

We used 32 mini-batches in the first stage and 4 in the second stage. In both training stages, we used stochastic gradient descent (SGD) optimization algorithm. We set the initial learning rate to $10^{-4}$ and applied cosine learning rate decay method until the loss converged. As performance indicators for the final model, we used Pearson correlation coefficient (PLCC), Spearman rank order correlation coefficient (SRCC) and root mean squared error (RMSE) to evaluate the accuracy of the predicted scores in comparison to the ground truth quality scores. These indicators are all widely used to assess the performance of IQA and PCQA models. All our experiments were run on a single NVIDIA RTX 3090 GPU.

\begin{table}[t]
\begin{center}
\caption{Ablation study of the COPP-Net model on the WPC database. AVE stands for average pooling, and CORA stands for correlation weight pooling strategy.}
\label{tab:ablation}
\setlength{\tabcolsep}{2.4mm}{
\begin{tabular}{|c|c|c|c|c|}
  \hline
  \multirow{2}{*}{Model}& \multirow{2}{*}{Pooling} &\multicolumn{3}{c|}{Criteria}   \\
  \cline{3-5}
    && PLCC & SRCC & RMSE  \\
  \hline
   
   PointNet++ SSG~\cite{qi2017pointnet++} & AVE &0.7136 &0.6988 &15.65 \\
   ARKP$_{t}$ w/o Stride Conv & AVE &0.7724 &0.7653 &14.86 \\
   ARKP$_{t}$ w/o R-Sampling & AVE &0.8461 &0.8286 &13.74 \\
   ARKP$_{t}$ w/o KNN & AVE & 0.7538 & 0.7346 &  15.03\\
   ARKP$_{t}$& AVE &0.8589 &0.8446 &11.75 \\
   ARKP$_{s}$& AVE &0.8691 &0.8616 &10.91 \\
   ARKP$_{t}$+ARKP$_{s}$ & AVE & 0.9015 & 0.8994 & 9.92 \\
   ARKP$_{t}$+ARKP$_{s}$ & CORA &\textbf{0.9324}&\textbf{0.9251}&\textbf{8.10}\\
   
  \hline
\end{tabular}}
\end{center}
\end{table}

\subsection{Prediction performance on WPC database}
Waterloo Point Cloud Database (WPC) includes 20 high-quality color point clouds, covering various geometric and texture characteristics. The database includes 37 distorted versions for each original point clouds, and there are in total 20 original point clouds and 740 distorted point clouds in the database. Each point cloud is annotated with their ground truth quality scores, obtained by subjective testing. In our experiments, we used the same division into training and testing sets as described in~\cite{liu2021pqa}~\cite{zhang2022no}: the point clouds for $Banana$, $Cauliflower$, $Mushroom$, $Pineapple$, accounting for 20\% of the whole data set, were used as test set, and the remaining point clouds were used as training set.

Table~\ref{tab:wpc_contrast} lists the overall results of COPP-Net and the selected benchmark NR methods~\cite{liu2021pqa}~\cite{zhang2022no}~\cite{liu2022point}~\cite{yang2022no} on the WPC test set. The results show that the proposed COPP-Net outperforms the prior state-of-the-art NR methods with a large margin. 

In Table~\ref{tab:combat FR}, we summarize the performance of several FR methods~\cite{mekuria2016evaluation}~\cite{mekuria2017performance}~\cite{meynet2020pcqm}~\cite{wang2004image}~\cite{yang2020inferring}~\cite{sheikh2006image}~\cite{liu2022perceptual} on different test contents in the WPC database separately, as given in~\cite{liu2022perceptual}, as well as the results for the same contents with the state-of-the-art NR models~\cite{liu2021pqa}~\cite{zhang2022no}~\cite{liu2022point}~\cite{yang2022no}, and the proposed COPP-Net obtained from our experiments. According to the results, COPP-Net is not only superior to the best performing prior NR methods in each category, but also outperforms the strongest FR method on all source contents. In general, FR models perform better than NR models, because they use the original pristine point cloud as a reference.

\subsection{Ablation Study}
To validate the efficacy of the proposed ARKP and CORA networks, we performed experiments on the WPC database using COPP-Net with various network configurations and pooling strategies. The results presented in Table~\ref{tab:ablation} demonstrate the effectiveness of the modifications we introduced in the ARKP networks. Specifically, the ARKP$_t$ and ARKP$_s$ models exhibit superior feature generation capabilities in comparison to the baseline PointNet++ SSG. Our analysis reveals that the Adaptive Stride Convolution and KNN components play a significant role in boosting performance, whereas the R-sampling component has a minor impact on accuracy, but it contributes to increasing the training speed.

Table~\ref{tab:ablation} demonstrates that the combination of ARKP$_t$ and ARKP$_s$ outperforms the individual networks. This indicates that the features generated by these two networks capture different aspects of quality, highlighting the importance of using distinct features for texture and structure. Furthermore, the use of weighted average pooling instead of average pooling in the CORA network enhances the performance, indicating that the model successfully leverages the quality imbalances across various regions of the point cloud.

\begin{table*}[t]       %Cross database comparison
\begin{center}
\footnotesize
\caption{Experiments on other databases. We compared the COPP-Net with the best performing prior NR models. All models use their default configurations and they are trained from scratch.} \label{tab:Cross database}
\setlength{\tabcolsep}{2.5mm}{
\begin{tabular}{|c|ccc|ccc|ccc|ccc|}
	\hline
	 \multirow{2}{*}{Method} & \multicolumn{3}{c|}{SJTU-PCQA} & \multicolumn{3}{c|}{SIAT-PCQD} & \multicolumn{3}{c|}{WPC2.0}& \multicolumn{3}{c|}{LS-PCQA Part I} \\
	\cline{2-13}       
	  & PLCC & SRCC & RMSE & PLCC & SRCC & RMSE & PLCC & SRCC & RMSE & PLCC & SRCC & RMSE\\
	\hline

    NR-3DQA~\cite{zhang2022no}
    &0.5987&0.5676&2.15&0.0327&0.0148&0.13&0.3944&0.4303&23.31&0.3861&0.3634&0.84 \\
    ResSCNN~\cite{liu2022point}
    &\textbf{0.9261}&\textbf{0.9099}&1.49&0.3083&0.2613&0.11&0.8213&0.8208&15.19&0.4131&0.4085&0.70 \\
    IT-PCQA~\cite{yang2022no} 
    &0.5922&0.5269&2.09&0.4266&0.0824&0.10&0.4571&0.3863&20.45&0.3159&0.3023&0.78 \\
    Our COPP-Net  
    &0.9257&0.8915&\textbf{1.31}&\textbf{0.8512}&\textbf{0.7115}&\textbf{0.08}&\textbf{0.8685}&\textbf{0.8706}&\textbf{12.78}&\textbf{0.5885}&\textbf{0.5949}&\textbf{0.68} \\
 	\hline
\end{tabular}}
\medskip
\end{center}
\end{table*}

\subsection{Prediction performance on other databases}
To provide a more comprehensive evaluation of the proposed model, we conducted comparative studies using COPP-Net and three state-of-the-art NR methods on five additional publicly available point cloud quality databases, including SJTU-PCQA~\cite{yang2020predicting}, SIAT-PCQD~\cite{wu2021subjective}, WPC2.0~\cite{liu2021reduced}, and LS-PCQA parts I~\cite{liu2022point}. In each dataset, we adopted an 80\%-20\% split for training and testing, respectively. PQA-Net~\cite{liu2021pqa} is not included, because it cannot be used on databases with multiple distortion types.

As shown in Table~\ref{tab:Cross database}, our COPP-Net outperforms most of the other methods across multiple databases, and even in cases where it does not achieve the best performance, the difference against the best performing method is typically small. These results highlight that the proposed method is widely applicable on different databases.

\section{Conclusions}
In this paper, we proposed a novel method for NR point cloud quality assessment, named as COPP-Net. The method takes into account the impact of different quality levels in different parts of the point cloud to the overall quality by using CORA network to constrain the dispersion of local quality levels. The ARKP network used in COPP-Net shows stronger feature generation capability than the baseline, demonstrating the importance of generating both texture and structure features of the point cloud. The experimental results show that the proposed method outperforms state-of-the-art FR and NR methods for PCQA.

The main limitation of COPP-Net is its limited scalability when increasing the number of patches: the CORA network needs to calculate all F$_{patch}$ features of a point cloud in parallel, which requires a significant amount of GPU memory. In contrast, the proposed ARKP network requires much less GPU memory, and the ARKP network without CORA can also achieve satisfactory performance.

\bibliographystyle{IEEEtran}        % 引用格式的文件名
\bibliography{IEEEexample}          % 引用的具体文件

% \vspace{12pt}
% \color{red}
% IEEE conference templates contain guidance text for composing and formatting conference papers. Please ensure that all template text is removed from your conference paper prior to submission to the conference. Failure to remove the template text from your paper may result in your paper not being published.

\end{document}